\documentclass[11pt,a4paper]{article}
\usepackage{times,latexsym}
\usepackage{url}
\usepackage[T1]{fontenc}

\usepackage[acceptedWithA]{tacl2018v2}
\usepackage{xspace,mfirstuc,tabulary}

\newif\iftaclinstructions
\taclinstructionsfalse % AUTHORS: do NOT set this to true
\iftaclinstructions
\renewcommand{\confidential}{}
\renewcommand{\anonsubtext}{(No author info supplied here, for consistency with
TACL-submission anonymization requirements)}
\newcommand{\instr}
\fi

\iftaclpubformat % this "if" is set by the choice of options

\else

\fi

\usepackage{tabularx}
\usepackage{graphicx}
\usepackage{amsmath}
\usepackage{bm}
\usepackage{booktabs}
\usepackage{algorithm}
\usepackage{algpseudocode}

\newcolumntype{C}{>{\centering\arraybackslash}X}
\newcolumntype{L}{>{\raggedright\arraybackslash}X}
\newcolumntype{R}{>{\raggedleft\arraybackslash}X}

\AtBeginDocument{
  \intextsep=0.4\intextsep
  \textfloatsep=0.4\textfloatsep}

\title{Unsupervised Abstractive Opinion Summarization \\ by Generating Sentences with Tree-Structured Topic Guidance}

\author{
 Masaru Isonuma$^{1}$ \quad Junichiro Mori$^{1,2}$ \quad Danushka Bollegala$^{3}$ \quad Ichiro Sakata$^{1}$ \\
 $^1$ The University of Tokyo \quad $^2$ RIKEN \quad $^3$ University of Liverpool \\
  {\sf isonuma@ipr-ctr.t.u-tokyo.ac.jp \quad mori@mi.u-tokyo.ac.jp} \\ 
  {\sf danushka@liverpool.ac.uk \quad isakata@ipr-ctr.t.u-tokyo.ac.jp} \\
}

\date{}

\begin{document}
\maketitle
\begin{abstract}
This paper presents a novel unsupervised abstractive summarization method for opinionated texts.
While the basic variational autoencoder-based models assume a unimodal Gaussian prior for the latent code of sentences, we alternate it with a \emph{recursive Gaussian mixture}, where each mixture component corresponds to the latent code of a topic sentence and is mixed by a tree-structured topic distribution. 
By decoding each Gaussian component, we generate sentences with \emph{tree-structured topic guidance}, where the root sentence conveys generic content, and the leaf sentences describe specific topics.
Experimental results demonstrate that the generated topic sentences are appropriate as a summary of opinionated texts, which are more informative and cover more input contents than those generated by the recent unsupervised summarization model \cite{bravzinskas2020unsupervised}.
Furthermore, we demonstrate that the variance of latent Gaussians represents the granularity of sentences, analogous to Gaussian word embedding \cite{vilnis2015word}.
\end{abstract}

\begin{figure*}[t!]
\centering
\includegraphics[width=\textwidth]{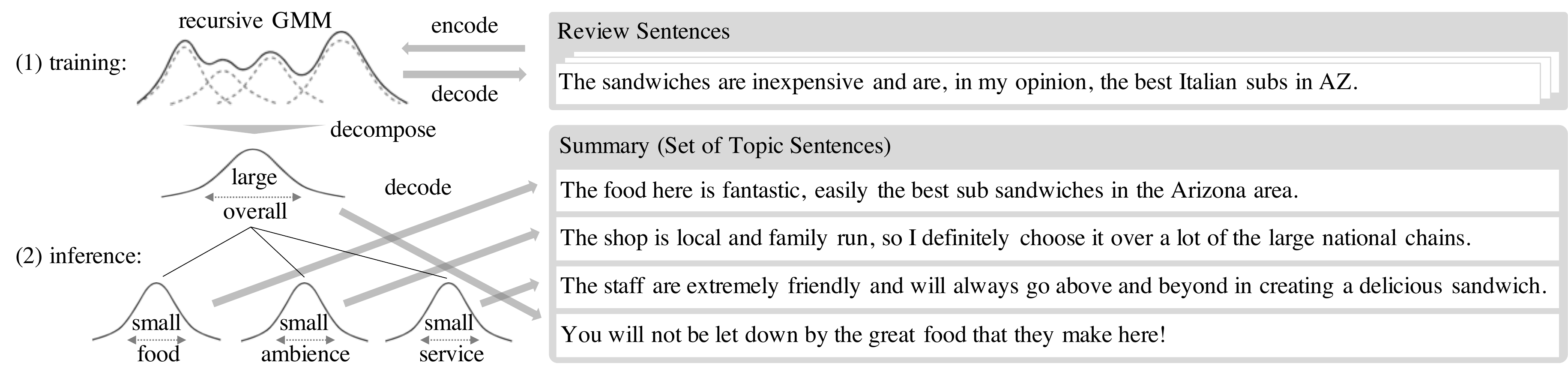}
\vspace{-1.5\baselineskip}
\caption{Outline of our approach. (1) The latent distribution of review sentences is represented as a recursive GMM and trained in an autoencoding manner. Then, (2) the topic sentences are inferred by decoding each Gaussian component. An example of a restaurant review and its corresponding gold summary are displayed.}
\vspace{-\baselineskip}
\label{fig:introduction}
\end{figure*}

\section{Introduction}
\label{sec:introduction}

Summarizing opinionated texts, such as product reviews and online posts on websites, has attracted considerable attention recently along with the development of e-commerce and social media.
Although extractive approaches are widely used in document summarization \cite{erkan2004lexpagerank, ganesan2010opinosis}, they often fail to provide an overview of the documents, particularly for opinionated texts \cite{carenini2013multi, gerani2014abstractive}.
Abstractive summarization can overcome this challenge by paraphrasing and generalizing an entire document. 
Although supervised approaches have seen significant success with the development of neural architectures \cite{see2017get, fabbri2019multi}, they are limited to specific domains, e.g., news articles, where a large number of gold summaries are available.
However, the domain of opinionated texts is diverse; manually writing gold summaries is therefore costly.

This lack in gold summaries has motivated prior work to develop unsupervised abstractive summarization of opinionated texts, e.g., product reviews \cite{chu2019meansum, bravzinskas2020unsupervised, amplayo2020unsupervised}.
While they generated consensus opinions by condensing input reviews, two key components were absent: \emph{topics} and \emph{granularity}, i.e., the level of detail.
For instance, as shown in \autoref{fig:introduction}, a gold summary of a restaurant review provides the overall impression and details about certain topics, such as food, ambience, and service.
Hence, a summary typically comprises diverse topics, some of which are described in detail, whereas others are mentioned concisely.

From this investigation, we capture the \emph{topic-tree structure} of reviews and generate \emph{topic sentences}, i.e., sentences summarizing specified topics.
In the topic-tree structure, the root sentence conveys generic content, whereas the leaf sentences mention specific topics.
From the generated topic sentences, we extract sentences with appropriate topics and levels of granularity as a summary.
Regarding extractive summarization, capturing topics \cite{titov2008joint, isonuma2017extractive, angelidis2018summarizing} and topic-tree structure \cite{celikyilmaz2010hybrid, celikyilmaz2011discovery} is useful for detecting salient sentences.
To the best of our knowledge, this is the first study to use the topic-tree structure in unsupervised abstractive summarization.

The difficulty of generating sentences with tree-structured topic guidance lies in controlling the granularity of topic sentences.
\citet{wang2019topic} generated a sentence with designated topic guidance, assuming that the latent code of an input sentence can be represented by a Gaussian mixture model (\textbf{GMM}), where each Gaussian component corresponds to the latent code of a topic sentence.
While they successfully generated a sentence relating to a designated topic by decoding each mixture component, modelling the sentence granularity in a latent space to generate topic sentences with multiple granularities remains to be realized.

To overcome this challenge, we model the sentence granularity by the variance size of the latent code.
We assume that general sentences have more uncertainty and are generated from a latent distribution with a larger variance, analogous to Gaussian word embedding \cite{vilnis2015word}.
Based on this assumption, we represent the latent code of topic sentences with Gaussian distributions, where the parent Gaussian receives a larger variance and represents a more generic topic sentence than its children, as shown in \autoref{fig:introduction}.
To obtain the latent code characterized above, we introduce a \emph{recursive Gaussian mixture} prior to modelling the latent code of input sentences in reviews.
A recursive GMM consists of Gaussian components that correspond to the nodes of the topic-tree, and the child priors are set to the inferred parent posterior.
Because of this configuration, the Gaussian distribution of higher topics receives a larger variance and conveys more general content than lower topics.

The contributions of our work are as follows:

\vspace{-0.4\baselineskip}
\begin{itemize}
    \setlength{\leftskip}{-10pt} % 左余白
    \setlength{\parskip}{0pt} % 段落間
    \setlength{\itemsep}{0pt} % 項目間
    \item We propose a novel unsupervised abstractive opinion summarization method by generating sentences with tree-structured topic guidance.
    \item To model the sentence granularity in a latent space, we specify a Gaussian distribution as the latent code of a sentence and demonstrate that the granularity depends on the variance size.
    \item Experiments demonstrate that the generated summaries are more informative and cover more input content than the recent unsupervised summarization \cite{bravzinskas2020unsupervised}.
\end{itemize}

\begin{figure*}[t!]
\centering
\vspace{-\baselineskip}
\includegraphics[width=\textwidth]{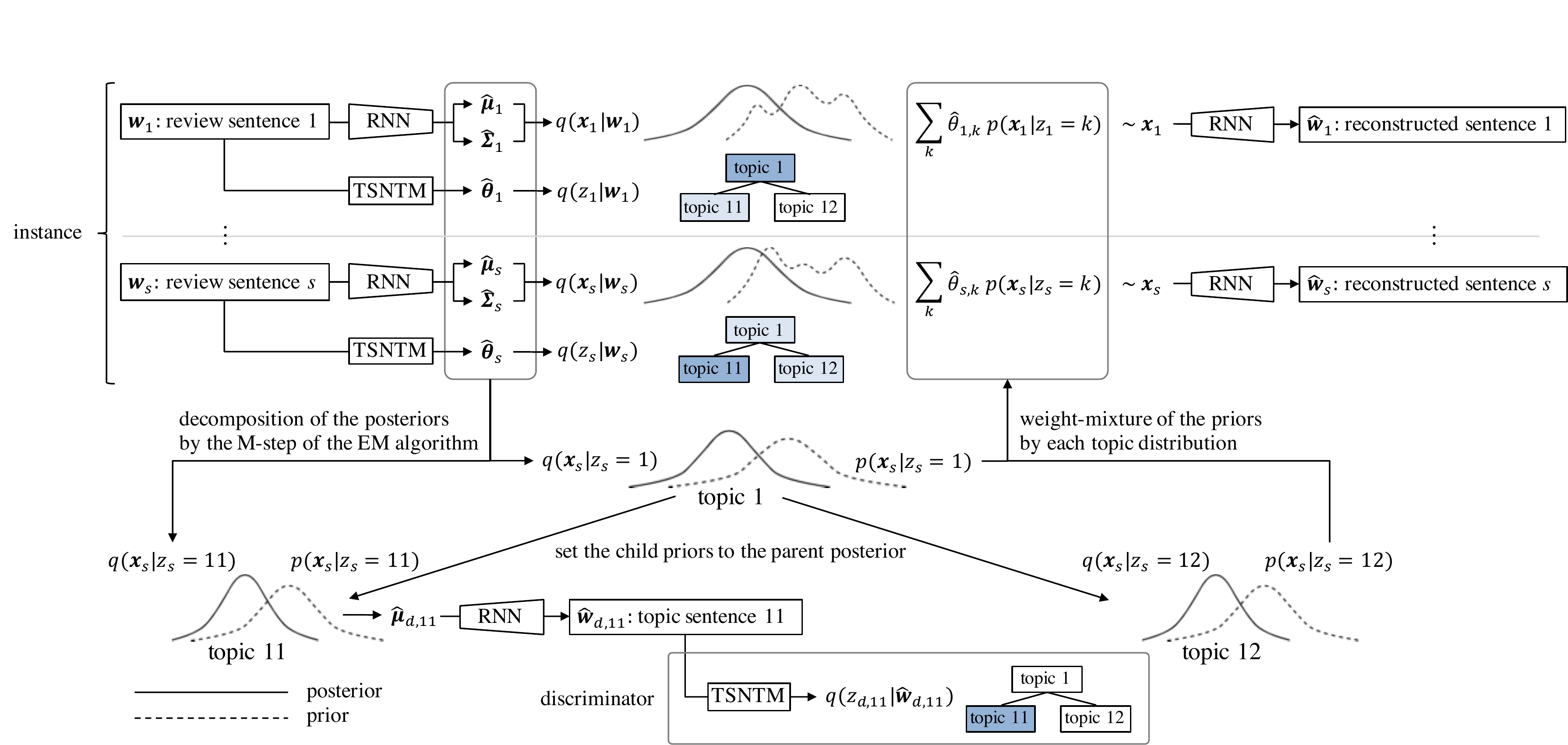}
\vspace{-1.5\baselineskip}
\caption{Outline of our model. We set a recursive Gaussian mixture as the latent prior of review sentences and obtain the latent posteriors of topic sentences by decomposing the posteriors of review sentences.}
\vspace{-0.5\baselineskip}
\label{fig:outline}
\end{figure*}

\section{Preliminaries}

\citet{bowman2016generating} adapted the variational autoencoder \cite[\textbf{VAE};][]{kingma2014auto, rezende2014stochastic} to obtain the density-based latent code of sentences. 
They assume the generative process of documents to be as follows:
\begin{align}
&\text{For each document index $d \!\in\! \{1,\ldots,D\}$:} \nonumber \\
&\hspace{\parindent}\text{For each sentence index $s \!\in\! \{1,\ldots,S_d\}$ in $d$:} \nonumber \\
&\hspace{\parindent}\text{1. Draw a latent code of the sentence $\bm{x}_s \!\in\! \mathcal{R}^n$:} \nonumber \\
&\hspace{2\parindent} {\textstyle \bm{x}_s \sim p(\bm{x}_s)} \label{eq:gaussian} \\
&\hspace{\parindent}\text{2. Draw a sentence $\bm{w}_s$:} \nonumber \\
&\hspace{2\parindent} {\textstyle \bm{w}_s|\bm{x}_s \!\sim\! p(\bm{w}_s|\bm{x}_s)} \!=\! \mathrm{RNN}(\bm{x}_s)
\end{align}
where $p(\bm{w}_s|\bm{x}_s) \!=\! \prod_t p(w_s^t|\bm{w}_s^{<t}, \bm{x}_s)$ is derived by an recurrent neural networks (\textbf{RNN}) decoder.
The latent prior is a standard Gaussian: $p(\bm{x}_s) \!=\! \mathcal{N}(\bm{x}_s|\bm{\mu}_0, \bm{\Sigma}_0)$.
The likelihood of a document and its evidence lower bound (\textbf{ELBO}) are given by \eqref{eq:likelihood_pre} and \eqref{eq:elbo_pre}, respectively:

{\footnotesize
\vspace{-\baselineskip}
\begin{align}
&p(\bm{W}_{1:S_d}) \!=\! \prod_{s=1}^{S_d} \Bigl\{\int\! p(\bm{w}_s|\bm{x}_s) p(\bm{x}_s) d\bm{x}_s \Bigr\} \label{eq:likelihood_pre}  \\
&\mathcal{L}_d \!=\! \sum_{s=1}^{S_d} \Bigl\{\! \mathbf{E}_{q(\bm{x}_s|\bm{w}_s)} \bigl[\log p(\bm{w}_s|\bm{x}_s) \bigr] \!-\! \mathrm{D_{KL}}\bigl[q(\bm{x}_s|\bm{w}_s)|p(\bm{x}_s)\bigr] \Bigr\} \label{eq:elbo_pre}
\end{align}
}%
$q(\bm{x}_s|\bm{w}_s) \!=\! \mathcal{N}(\bm{x}_s|\hat{\bm{\mu}}_s, \hat{\bm{\Sigma}}_s)$ is the variational distribution with $\hat{\bm{\mu}}_s\!=\!f_{\mu}(\bm{w}_s), \hat{\bm{\Sigma}}_s\!=\!\mathrm{diag}[f_{\Sigma}(\bm{w}_s)]$ where $f_{\mu}$ and $f_{\Sigma}$ are RNN encoders.

% Extending their work, \citet{wang2019topic} constructed the prior as a GMM: $p(\bm{x}_s) \!=\! \sum_k p(z_s\!=\!k) p(\bm{x}_s|z_s\!=\!k) \!=\! \sum_{k} p(z_s\!=\!k) \mathcal{N}(\bm{x}_s|\bm{\mu}_k, \bm{\Sigma}_k)$ where $\mathcal{N}(\bm{x}_s|\bm{\mu}_k, \bm{\Sigma}_k)$ denotes the latent distribution of a topic sentence specified by topic $k$. 
% A topic sentence is obtained from $\mathcal{N}(\bm{x}_s|\bm{\mu}_k, \bm{\Sigma}_k)$.

By representing sentences by Gaussians rather than vectors, the decoded sentence from the intermediate latent code between two sentences is grammatical and has a coherent topic with the two sentences.
Extending their work, we construct the prior as a recursive GMM and infer the topic sentences by decoding each Gaussian component.

\section{RecurSum: Recursive Summarization}

In this section, we explain our model, RecurSum. 
\autoref{fig:outline} shows the outline.
The latent code of review sentences is obtained as a recursive GMM (\ref{sec:documentgeneration}), and topic sentences are inferred by decoding each Gaussian component (\ref{sec:inference}).
A summary is then created by extracting the appropriate topic sentences (\ref{sec:extraction}).
We introduce additional components to improve the quality of topic sentences (\ref{sec:addition}) and explain why general/specific content is conveyed by the root/leaf topics, referring to the analogy with Gaussian word embedding (\ref{sec:word2gauss}).

\subsection{Generative Model of Reviews}
\label{sec:documentgeneration}

We assume the generative process of reviews to be as follows. 
We refer to the set of sentences in multiple reviews of a specific product as \emph{instance}.
Compared to \citet{bowman2016generating}, we explicitly model the topic of review sentences as follows:
\begin{align}
&\text{For each instance index $d \!\in\! \{1,\ldots,D\}$:} \nonumber \\
&\hspace{\parindent}\text{For each sentence index $s \!\in\! \{1,\ldots,S_d\}$ in $d$:} \nonumber \\
&\hspace{\parindent}\text{1. Draw a topic of the sentence $z_s \!\in\! \{1, \ldots, K\}$:} \nonumber \\
&\hspace{2\parindent} {\textstyle z_s \sim \mathrm{Mult}(\bm{\theta})} \\
&\hspace{\parindent}\text{2. Draw a latent code of the sentence $\bm{x}_s \!\in\! \mathcal{R}^n$:} \nonumber \\
&\hspace{2\parindent} {\textstyle \bm{x}_s|z_s \sim \prod_{k=1}^{K} p(\bm{x}_s|z_s\!=\!k)^{\delta(z_s=k)}}  \label{eq:recursivegmm} \\
&\hspace{\parindent}\text{3. Draw a review sentence $\bm{w}_s$:} \nonumber \\
&\hspace{2\parindent} {\textstyle \bm{w}_s|\bm{x}_s \!\sim\! p(\bm{w}_s|\bm{x}_s)} \!=\! \mathrm{RNN}(\bm{x}_s)
\end{align}
where the topic distribution is tree-structured, and its prior is set to be uniform.
In \eqref{eq:recursivegmm}, we assume a recursive GMM as the latent prior of a review sentence ($\delta$ is a Dirac delta).
Each mixture component corresponds to the latent distribution of a sentence conditioned on a specific topic, $p(\bm{x}_s|z_s\!=\!k)$:

{\footnotesize
\vspace{-\baselineskip}
\begin{align}
p(\bm{x_s}|z_s\!=\!1) &= \mathcal{N}(\bm{x}_s|\bm{\mu}_0, \bm{\Sigma}_0) \label{eq:prior_root} \\
\begin{split}
p(\bm{x_s}|z_s\!=\!k) &= q(\bm{x_s}|z_s\!=\!par(k)) \\
&= \mathcal{N}(\bm{x}_s|\hat{\bm{\mu}}_{d,par(k)}, \hat{\bm{\Sigma}}_{d,par(k)}) \ (k \neq 1) \label{eq:prior_topic}
\end{split}
\end{align}
}%
where $par(k)$ denotes the parent of the $k$-th topic.
$q(\bm{x_s}|z_s\!=\!par(k))$ is the approximated latent posterior of the parent topic sentence as derived later in \autoref{sec:inference}.
We assume that the latent posterior of the parent sentence is appropriate as the latent prior of its child sentences.
% We specify it as the latent prior of the $k$-th topic sentence.

Under our generative model, the likelihood of an instance and its ELBO are given by \eqref{eq:likelihood} and \eqref{eq:elbo}, respectively:

{\footnotesize
\vspace{-\baselineskip}
\begin{align}
p(\bm{W}_{1:S_d}) \!=\! \prod_{s=1}^{S_d} \int\! p(\bm{w}_s|\bm{x}_s) p(\bm{x}_s|z_s) p(z_s) d\bm{x}_s dz_s \label{eq:likelihood} 
\end{align}
\vspace{-\baselineskip}
\begin{align}
\begin{split}
\mathcal{L}_d \!=\!& \sum_{s=1}^{S_d} \Bigl\{
    \mathbf{E}_{q(\bm{x}_s|\bm{w}_s)} \bigl[\log p(\bm{w}_s|\bm{x}_s) \bigr] \\
    & \!-\! \mathbf{E}_{q(\bm{x}_s|\bm{w}_s)q(z_s|\bm{w}_s)} \bigl[\log q(\bm{x}_s|\bm{w}_s) \!-\! \log p(\bm{x}_s|z_s) \bigr] \\
    & \!-\! \mathbf{E}_{q(z_s|\bm{w}_s)} \bigl[\log q(z_s|\bm{w}_s) \!-\! \log p(z_s) \bigr]
\Bigr\} \\ 
\!=\!& \sum_{s=1}^{S_d} \Bigl\{ \mathbf{E}_{q(\bm{x}_s|\bm{w}_s)} \bigl[\log p(\bm{w}_s|\bm{x}_s) \bigr] \!-\! \mathrm{D_{KL}}\bigl[q(z_s|\bm{w}_s)|p(z_s) \bigr] \Bigr\} \\
    & \!-\! \sum_{k=1}^{K} \sum_{s=1}^{S_d} \Bigl\{ \hat{\theta}_{s,k} \mathrm{D_{KL}}\bigl[q(\bm{x}_s|\bm{w}_s)|p(\bm{x}_s|z_s\!=\!k) \bigr] \Bigr\} \label{eq:elbo}
\end{split}
\end{align}
}%
where $q(\bm{x}_s|\bm{w}_s) \!=\! \mathcal{N}(\bm{x}_s|\hat{\bm{\mu}}_s, \hat{\bm{\Sigma}}_s)$ is the latent posterior of a sentence $s$, inferred by an RNN encoder.
$\hat{\theta}_{s,k}\!=\!q(z_s\!=\!k|\bm{w}_s)$ is the variational topic distribution and inferred by the tree-structured neural topic model \cite[\textbf{TSNTM};][]{isonuma2020tree}. 
More details are provided in Appendix \ref{app:inferencetopic}.

\subsection{Inference of Topic Sentences}
\label{sec:inference}

From the latent posterior of review sentences, we infer the latent posterior of each topic sentence using the M-step of the EM algorithm.
We define the variational distribution of the latent code of a topic sentence as \eqref{eq:em} and compute the Gaussian parameters as \eqref{eq:mu_k} and \eqref{eq:Sigma_k} that maximize $\sum_{s=1}^{S_d} \mathbf{E}_{q(\bm{x}_s|\bm{w}_s) q(z_s|\bm{w}_s)} \bigl[ \log q(\bm{x}_s|z_s) \bigr]$ as follows:

{\footnotesize
\vspace{-\baselineskip}
\begin{align}
& {\textstyle q(\bm{x}_s|z_s) \!=\! \prod_{k=1}^{K} \mathcal{N}(\bm{x}_s|\hat{\bm{\mu}}_{d,k}, 
\hat{\bm{\Sigma}}_{d,k})^{\delta(z_s=k)}} \label{eq:em} \\
\hat{\bm{\mu}}_{d,k} & \!=\! \frac{\sum_{s=1}^{S_d} \hat{\theta}_{s,k} \mathbf{E}_{q(\bm{x}_s|\bm{w}_s)}[\bm{x}_s]}{\sum_{s=1}^{S_d} \hat{\theta}_{s,k}}
\!=\! \frac{\sum_{s=1}^{S_d} \hat{\theta}_{s,k} \hat{\bm{\mu}}_s}{\sum_{s=1}^{S_d} \hat{\theta}_{s,k}} \label{eq:mu_k}\\
\hat{\bm{\Sigma}}_{d,k} & \!=\! \frac{\sum_{s=1}^{S_d} \hat{\theta}_{s,k} \mathbf{E}_{q(\bm{x}_s|\bm{w}_s)}[(\bm{x}_s\!-\!\hat{\bm{\mu}}_{d,k})(\bm{x}_s\!-\!\hat{\bm{\mu}}_{d,k})^{\!\top}]}{\sum_{s=1}^{S_d} \hat{\theta}_{s,k}} \nonumber \\
& \!=\! \frac{\sum_{s=1}^{S_d} \hat{\theta}_{s,k} \{ \hat{\bm{\Sigma}}_s\!+\!(\hat{\bm{\mu}}_s\!-\!\hat{\bm{\mu}}_{d,k})(\hat{\bm{\mu}}_s\!-\!\hat{\bm{\mu}}_{d,k})^{\!\top} \}}{\sum_{s=1}^{S_d} \hat{\theta}_{s,k}} \label{eq:Sigma_k}
\end{align}
}%
From these latent posteriors, we generate the topic sentences for each instance using the respective mean not a sample: $\hat{\bm{w}}_{d,k} \!\sim\! p(\bm{w}_{d,k}|\hat{\bm{\mu}}_{d,k}) \!=\! \mathrm{RNN}(\hat{\bm{\mu}}_{d,k})$.
Similar to \citet{bravzinskas2020unsupervised, chu2019meansum}, we assume that the average latent code represents the common contents of the corresponding topic, while specific contents are distributed apart from the mean.
Therefore, decoding the mean rather than a sample would be desirable for generating a summary.

\subsection{Extraction of Summary Sentences}
\label{sec:extraction}

Next, we create a summary by extracting appropriate sentences from the generated topic sentences.
As gold summaries are not available for training, we need a measure to evaluate candidate summaries using only input reviews.
As reported in \citet{chu2019meansum}, the ROUGE scores \cite{lin2004rouge} between a candidate summary and the \emph{input reviews} effectively measures the extent to which the summary encapsulates the reviews.
Based on this assumption, we search the topic sentences by maximizing the ROUGE-1 F-measure with the review sentences in an instance.
We use a beam search and keep multiple highest-score candidates for each step.
Similar to \citet{carbonell1998use}, to eliminate the redundancy of summary sentences, we do not add a sentence with a high word overlap (ROUGE-1 precision) against the sentences already included in the summary.
The hyperparameters are tuned based on the validation set, as described in \autoref{sec:implementation}.

After selecting the summary sentences, we sort them in the depth-first order according to the topic-tree structure, i.e., we begin at the root node and explore as far as possible along each branch before backtracking.
\citet{barzilay2008modeling} advocate that adjacent sentences in the coherent text tend to have similar contents.
As we assume that sentences linked by parent-child relations are topically coherent, the generated summary is expected to be locally coherent by extracting child sentences after their parent sentence.

\subsection{Additional Model Components}
\label{sec:addition}

The basic components of our model have been explained in the previous sections.
This section introduces three additional components to improve the quality of topic sentences.
In ablation studies (\autoref{sec:ablation}), we will see the effect of these components on summarization performance.

\paragraph{Discriminator}

% To prevent the parent-child topics from becoming too close to differentiate, we introduce a discriminator to ensure that each topic sentence has a specific topic, drawing inspiration from \citet{hu2017toward, tang2019topic}.
To ensure that each topic sentence has a specific topic, we introduce a discriminator following \citet{hu2017toward, tang2019topic}.
We approximate the sample of the topic sentence by using the Gumbel-softmax trick \cite{jang2017categorical, maddison2017concrete} and reuse the TSNTM to estimate the topic distribution of the sample, $q(z_{d,k}|\hat{\bm{w}}_{d,k})$.
By maximizing the likelihood of the specified topic as \eqref{eq:discriminator}, the discriminator forces the generated $k$-th topic sentence to be coherent with topic $k$.

{\small
\vspace{-\baselineskip}
\begin{align}
{\textstyle \mathcal{L}_d^{disc} \!=\! \sum_{k=1}^{K} \log q(z_{d,k}\!=\!k|\hat{\bm{w}}_{d,k})}
\label{eq:discriminator}
\end{align}
}%

\paragraph{Attention}

We use the attention-based RNN decoder \cite{luong2015effective} to efficiently reflect input sentence information into output topic sentences.
Given the hidden state of the $t$-th word in an output sentence $\bm{h}_o^t$ and the $i$-th word in an input review sentence $\bm{h}_s^{i}$, we calculate the attention distribution over all the words in the input review sentences to compute the word probability.

{\small
\vspace{-\baselineskip}
\begin{align}
a(\bm{h}_o^t, \bm{h}_s^{i}) &= \frac{\exp(\bm{h}_o^{t\top} \bm{h}_s^{i})}{\sum_{s'}\sum_{i'} \exp(\bm{h}_o^{t\top} \bm{h}_{s'}^{i'})} \\
\bm{c}_o^t &= {\textstyle \sum_{s'} \sum_{i'} a(\bm{h}_o^t, \bm{h}_{s'}^{i'}) \bm{h}_{s'}^{i'}} \\
p(w_o^t|\bm{w}_o^{<t}, \hat{\bm{\mu}}_o) &= \mathrm{softmax}(\bm{W}[\bm{h}_o^t; \bm{c}_o^t])
\end{align}
}%

\paragraph{Nucleus Sampling}

During the inference, we use nucleus sampling \cite{holtzman2019curious} to decode the topic sentences.
They reported that maximization-based decoding methods such as beam search tend to generate bland, incoherent, and repetitive text in open-ended text generation.
As we will see in the ablation experiments, nucleus sampling is effective in generating diverse and informative topic sentences.

\begin{figure}[t!]
\centering
\includegraphics[width=7.4cm]{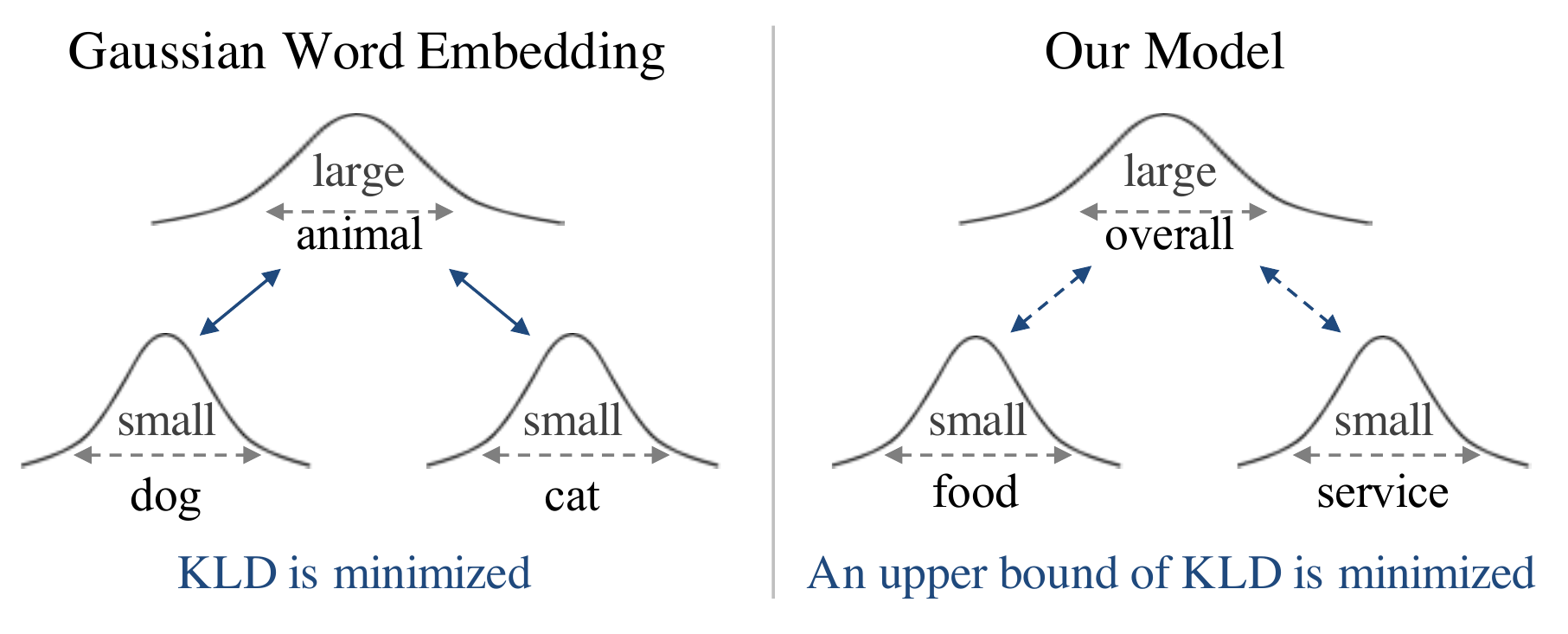}
\vspace{-\baselineskip}
\caption{Analogy with Gaussian word embedding.}
\label{fig:word2gauss}
\end{figure}

\subsection{Analogy with Gaussian Word Embedding}
\label{sec:word2gauss}

Here, we explain why a general sentence is generated from the root topic, while more specific content is conveyed by the sentences generated by the leaf topics, referring to Gaussian word embedding.

Gaussian word embedding \cite{vilnis2015word} represents words as Gaussian distributions and captures the hierarchical relations among the words.
As shown in \autoref{fig:word2gauss}, by representing words as densities over a latent space and minimizing the KL-divergence of the distributions, they detect that common words such as ``animal'' obtain a larger variance than more specific words, such as ``dog'' and ``cat''.
This can be explained by the fact that general words have more uncertainty in their meaning (i.e.,``animal'' sometimes denotes ``dog'' and other times ``cat'').

Similarly, our model minimizes the upper bound of the KL-divergence of the latent distribution between a parent topic sentence and its children.
In \eqref{eq:kl_geq}, we show that the $x$-related term in the ELBO \eqref{eq:elbo} is an upper bound of the KL divergence of the latent posteriors between parent-child topic sentences (derived in Appendix \ref{app:proof}).

{\footnotesize
\vspace{-\baselineskip}
\begin{align}
\begin{split}
\label{eq:kl_geq}
&{\textstyle \sum_{s=1}^{S_d} \hat{\theta}_{s,k} \mathrm{D_{KL}}\bigl[q(\bm{x}_s|\bm{w}_s)|p(\bm{x}_s|z_s\!=\!k) \bigr]} \\
&{\textstyle \!\geq\! \sum_{s=1}^{S_d} \hat{\theta}_{s,k} \mathrm{D_{KL}}\bigl[q(\bm{x}_s|z_s\!=\!k)|p(\bm{x}_s|z_s\!=\!k) \bigr]} \\
&{\textstyle \!=\! \sum_{s=1}^{S_d} \hat{\theta}_{s,k} \mathrm{D_{KL}}\bigl[q(\bm{x}_s|z_s\!=\!k)|q(\bm{x}_s|z_s\!=\!par(k)) \bigr]}
\end{split}
\end{align}
}%
since $p(\bm{x}_s|z_s\!=\!k) \!=\! q(\bm{x}_s|z_s\!=\!par(k))$ as defined in \eqref{eq:prior_topic}.
Similar to Gaussian word embedding, maximizing the ELBO forces the latent distribution of a parent to be close to that of its children, and the parent receives a larger variance than its children.
This property ensures that the parent-child topics have a coherent topic, and more general content is conveyed by the root topic sentences.
Intuitively, a general sentence, such as ``I love this restaurant'', includes several topics, such as ``food'' and ``service'', and has a large uncertainty of semantics.
Thus, we assume that a generic sentence is represented by the mean of the latent distribution with a larger variance, whereas a more specific sentence is generated from the distribution with a smaller variance.

Similar to \citet{vilnis2015word}, we observed that the eigenvalues of the full covariance of topic sentences \eqref{eq:Sigma_k} become extremely small during training.
To maintain a reasonably sized and positive semi-definite covariance, we add a hard constraint to the diagonal covariance of the review sentences as $\hat{\bm{\Sigma}}_{s,ii} \!\leftarrow\! \max(\lambda, \hat{\bm{\Sigma}}_{s,ii})$ since $\log|\hat{\bm{\Sigma}}_{d,k}| \!\geq\! \ (\sum_{s} \hat{\theta}_{s,k} \log|\hat{\bm{\Sigma}}_s|) / (\sum_{s} \hat{\theta}_{s,k}) \!\geq\! n\log \lambda$, as derived in Appendix \ref{app:proof}.

\begin{table}[t!]
\footnotesize
\centering
\begin{tabularx}{\columnwidth}{p{1.6cm}RR}
\toprule
Dataset&Yelp&Amazon\\ 
\midrule
Training&173,088&280,692 \\ 
Validation&100&84 \\ 
Test&100&96 \\ 
\bottomrule
\end{tabularx}
\vspace{-0.5\baselineskip}
\caption{Number of instances (pairs of eight reviews and a gold summary) in the datasets. The training set does not contain gold summaries.}
\label{tbl:dataset}
\end{table}

\begin{table*}[t!]
\footnotesize
\begin{tabularx}{\textwidth}{p{5.5cm}RRR|RRR} 
\toprule
&\multicolumn{3}{c|}{Yelp}&\multicolumn{3}{c}{Amazon}\\ 
\midrule
Model&Rouge-1&Rouge-2&Rouge-L&Rouge-1&Rouge-2&Rouge-L \\ 
\midrule
Multi-Lead-1&27.42&3.74&14.34&30.32&\underline{5.85}&15.96 \\
LexRank \cite{erkan2004lexpagerank}&26.40&3.19&14.35&31.42&\underline{5.31}&16.70 \\
Opinosis \cite{ganesan2010opinosis}&25.80&2.92&14.57&28.90&4.11&16.33 \\
MeanSum \cite{chu2019meansum}&28.66&3.73&15.77&30.16&4.51&17.76 \\
Copycat \cite{bravzinskas2020unsupervised}&28.95&\underline{4.80}&\underline{17.76}&31.84&\underline{5.79}&\textbf{20.00} \\
DenoiseSum \cite{amplayo2020unsupervised}&29.77&\underline{5.02}&\underline{17.63}&--&--&-- \\
RecurSum (Our Model)&\textbf{33.24}&\textbf{5.15}&\textbf{18.01}&\textbf{34.91}&\textbf{6.33}&18.91 \\
\midrule
% RecurSum (Oracle)&35.59&7.93&28.63&37.17&9.85&30.19 \\
PEGASUS \cite{zhang2020pegasus}&\textit{32.03}&\textit{5.64}&\textit{18.08}&29.30&4.39&\textit{18.10} \\
\bottomrule
\end{tabularx}
\vspace{-0.5\baselineskip}
\caption{ROUGE F1-scores of the test set (\%). Boldface shows the highest score excluding the oracle, and underlined scores are not regarded as statistically significant ($p < 0.05$) by approximate randomization test as compared to the highest score.}
\vspace{-0.5\baselineskip}
\label{tbl:result}
\end{table*}

\section{Experiments}

\subsection{Datasets}
\vspace{-2pt}

In our experiments, we used the \textit{\textbf{Yelp} Dataset Challenge}\footnote{\url{https://www.yelp.com/dataset}} and \textit{\textbf{Amazon} product reviews} \cite{mcauley2015image}.
By pre-processing the reviews similarly as in \citet{chu2019meansum, bravzinskas2020unsupervised}, we obtained the dataset as shown in \autoref{tbl:dataset}.
Regarding the training set, we removed products\footnote{We refer to businesses (e.g., a specific Starbucks branch) in Yelp and products (e.g., iPhone X) in Amazon as \textit{products}.} fewer than $8$ reviews and reviews in which the maximum number of sentences exceeds $50$.
To prevent the dataset from being dominated by a small number of products, we created $12$ and $2$ instances for each product in Yelp and Amazon, respectively.
Then, we randomly selected $8$ reviews to construct an instance.
Regarding the validation/test set of Yelp, we randomly split 200 instances provided by \citet{chu2019meansum}\footnote{\url{https://github.com/sosuperic/MeanSum}} into validation and test sets.
For Amazon, we used the same validation and test sets provided by \citet{bravzinskas2020unsupervised}\footnote{\url{https://github.com/abrazinskas/Copycat-abstractive-opinion-summarizer}}.
These gold summaries were created by Amazon Mechanical Turk (\textbf{AMT}) workers, who summarized $8$ reviews for each product.
The vocabulary comprises words that appear more than $16$ times in the training set. 
The vocabulary sizes are $31,748$ and $30,732$ for Yelp and Amazon, respectively.

\subsection{Implementation Details}
\label{sec:implementation}

We set the hyperparameters as follows, which maximize the ROUGE-L in the validation set of Yelp and use the same hyperparameters on Amazon\footnote{\url{https://github.com/misonuma/recursum}}.
The dimensions of word embeddings and the latent code of the sentences are $200$ and $32$, respectively.
The encoder and decoder are single-layer bi-directional and uni-directional GRU-RNN \cite{chung2014empirical} with $200$-dimensional hidden units for each direction.
The threshold of nucleus sampling is $0.4$.
We train our model using Adam \cite{kingma2014adam} with a learning rate of $5.0\!\times\!10^{-3}$, a batch size of $8$, and a dropout rate of $0.2$.
The initial Gumbel-softmax temperature is set to $1$ and decreased by $2.5\!\times\!10^{-5}$ per training step.
Similar to \citet{bowman2016generating, yang2017improved}, we avoid posterior collapse by increasing the weight of the KL-term by $2.5\times10^{-5}$ per training step.
We set the review sentence's minimum covariance to $\lambda \!=\! \exp(0.5)$.
Regarding the tree structure, we set the number of levels to $3$, and the number of branches to $4$ for both the second and third levels. 
The total number of topics is $21$.
Regarding the summary sentence extractor in \autoref{sec:extraction}, we set the maximum number of extracted sentences as $6$, the beam width as $8$, and the redundancy threshold as $0.6$.

\subsection{Baseline Methods}
\label{sec:baseline}

As a baseline, we use \emph{Multi-Lead-1}, which extracts the first sentence of each review.
Furthermore, we employ unsupervised extractive approaches, \emph{LexRank} \cite{erkan2004lexpagerank} and \emph{Opinosis} \cite{ganesan2010opinosis}.
LexRank is a PageRank-based sentence extraction method that constructs a graph in which sentences and their similarity are represented by the nodes and edges, respectively.
Opinosis constructs a word-based graph and extracts redundant phrases as a summary. 
As unsupervised abstractive summarization methods, we use \emph{MeanSum} \cite{chu2019meansum}, \emph{Copycat} \cite{bravzinskas2020unsupervised} and \emph{DenoiseSum} \cite{amplayo2020unsupervised}.
MeanSum computes the mean of the review embeddings and decodes it as a summary.
Copycat generates a consensus opinion by a hierarchical VAE which is trained by generating a new review given a set of other reviews of a product.
DenoiseSum\footnote{As a complete code is not available, we report the result of different test splits from ours, which are used in their sample of output summaries. \url{https://github.com/rktamplayo/DenoiseSum}} creates synthetic reviews by adding noise to original reviews and generates a summary by removing non-salient information as noise.

As an upper bound of extraction methods, we also report the performance of \emph{Oracle}, which extracts the topic sentences such that they obtain the highest ROUGE-L against each gold summary.
As the average number of sentences in the gold summaries is approximately four, we extract four topic sentences to generate a summary.

\begin{table*}[t!]
\footnotesize
\centering
\begin{tabularx}{\textwidth}{p{0.8cm}RRRR|RRRR} \toprule
&\multicolumn{4}{c|}{Yelp}&\multicolumn{4}{c}{Amazon} \\ \midrule
Model&Fluency&Coherence&Informative.&Redundancy&Fluency&Coherence&Informative.&Redundancy \\ \midrule
LexRank&-16.88&-13.51&\underline{-0.64}&\underline{-6.83}&-18.18&-15.07&\underline{14.11}&\underline{-4.76} \\
MeanSum&\underline{5.63}&-16.18&-13.73&\underline{0.70}&\underline{2.74}&-14.69&-13.70&\underline{1.32} \\
Copycat&\textbf{15.07}&\underline{7.88}&-7.19&\textbf{4.00}&\textbf{14.65}&\underline{9.80}&-17.65&\textbf{6.85} \\
\emph{RecurSum}&\underline{-2.56}&\textbf{19.46}&\textbf{24.44}&\underline{2.78}&\underline{0.70}&\textbf{17.72}&\textbf{17.39}&\underline{-2.99} \\
\bottomrule
\end{tabularx}
\vspace{-0.5\baselineskip}
\caption{Human evaluation scores on the \textit{quality} of the summaries. 
The scores are computed by using the best-worst scaling (\%) and range from $-100$ (unanimously worst) to $+100$ (unanimously best).
Boldface denotes the highest score, and underlined scores are not regarded as statistically significant ($p<0.05$) by Tukey HSD test as compared to the highest score.}
\label{tbl:humaneval}
\vspace{-0.5\baselineskip}
\end{table*}

\begin{table*}[t!]
\footnotesize
\begin{minipage}[t]{0.45\textwidth}
\raggedleft
\begin{tabularx}{\textwidth}{p{0.5cm}RR|RR} \toprule
&\multicolumn{2}{c|}{Yelp}&\multicolumn{2}{c}{Amazon}\\ \midrule
&Copycat&\emph{RecurSum}&Copycat&\emph{RecurSum} \\ \midrule
Full&47.79&47.43&45.64&44.74 \\
Partial&41.59&40.00&40.94&38.95 \\
No&10.62&12.57&13.42&16.32 \\ \bottomrule
\end{tabularx}
\vspace{-0.5\baselineskip}
\caption{Human evaluation scores on the \textit{faithfulness} of the summaries (\%). 
The difference of each system's frequency distribution is not regarded as statistically significant ($p<0.05$) by $\chi^2$ test.}
\vspace{-0.5\baselineskip}
\label{tbl:faithfulness}
\end{minipage}
\hspace{0.02\textwidth}
\begin{minipage}[t]{0.53\textwidth}
\raggedright
\begin{tabularx}{\textwidth}{p{0.5cm}RRR|RRR} \toprule
&\multicolumn{3}{c|}{Yelp}&\multicolumn{3}{c}{Amazon}\\ \midrule
&Copycat&\emph{RecurSum}&Gold&Copycat&\emph{RecurSum}&Gold \\ \midrule
Full&23.94&\textbf{31.05}&34.52&29.15&\textbf{33.31}&38.41 \\
Partial&30.02&\textbf{37.73}&40.91&29.15&\textbf{36.53}&39.63 \\
No&46.04&\textbf{31.22}&24.58&41.71&\textbf{30.16}&21.96 \\ \bottomrule
\end{tabularx}
\vspace{-0.5\baselineskip}
\caption{Human evaluation scores on the \textit{coverage} of the summaries (\%).
The difference of the frequency distribution between Copycat and RecurSum is statistically significant ($p<0.05$) by $\chi^2$ test.}
\vspace{-0.5\baselineskip}
\label{tbl:coverage}
\end{minipage}
\end{table*}

\vspace{-0.3\baselineskip}
\subsection{Semi-automatic Evaluation of Summaries}

Following \citet{chu2019meansum, bravzinskas2020unsupervised}, we use the ROUGE-1/2/L F1-scores \cite{lin2004rouge} as semi-automatic evaluation metrics.

\autoref{tbl:result} shows the rouge scores of our model, \emph{RecurSum}, and the baselines for the test sets.
In most metrics on both datasets, our model outperforms MeanSum and achieves competitive performance compared with the recent unsupervised summarization model, Copycat. 
Regarding the oracle, our model significantly outperforms the other models.
This result suggests that our model can improve the performance by using more sophisticated extraction methods.
Although we have also attempted to use the integer linear programming-based method \cite{gillick2009scalable}, it did not improve the performance.
Developing such extraction techniques is beyond the scope of the current study, which focuses on topic structure and is deferred to future work.

\begin{figure*}[t!]
\centering
\includegraphics[width=16.0cm]{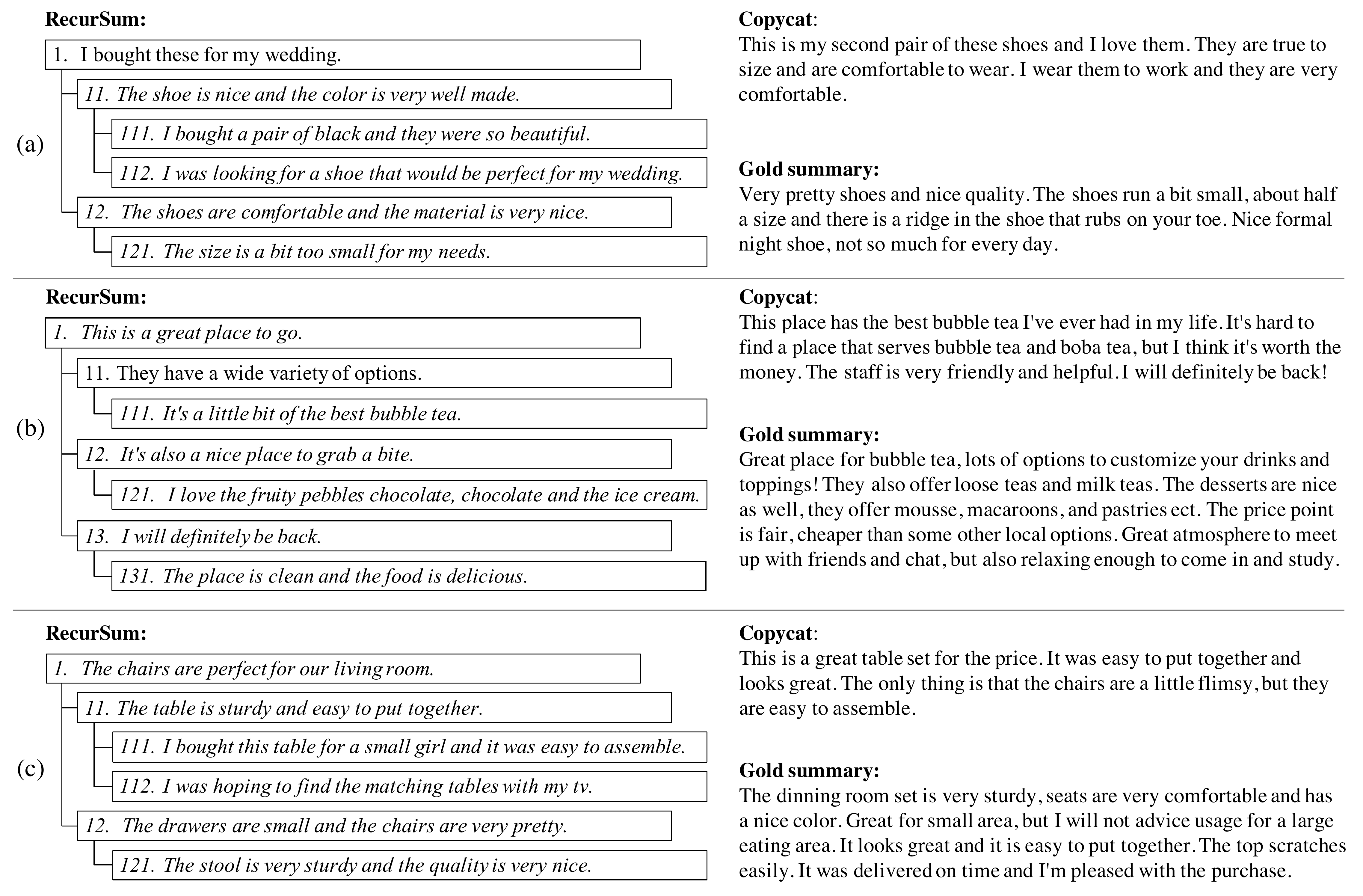}
\vspace{-1.5\baselineskip}
\caption{Generated topic sentences of (a) an Amazon review of heal shoes, (b) a Yelp review of a coffee shop, and (c) an Amazon review of table chair set. Topic sentences selected as a summary are highlighted in \textit{italic}.}
\vspace{-0.5\baselineskip}
\label{fig:sample}
\end{figure*}

\subsection{Human Evaluation of Summaries}
\label{sec:humanevaluation}

We conducted a human evaluation using AMT.
Following \citet{bravzinskas2020unsupervised, amplayo2020unsupervised}, we randomly selected $50$ instances from each test set and asked AMT workers\footnote{To obtain reliable answers, we set the worker requirements to 98\% approval rate, 1000+ accepted tasks, and locations in the US, UK, Canada, Australia, and New Zealand.} to answer the following three tasks:

\paragraph{Quality of the Summaries}

We presented four system summaries in random order and asked six AMT workers to rank the summzarization quality referring to the gold summary.
We compute each system's score as the percentage of times selected as the best minus those are selected as the worst by using the best-worst scaling \cite{louviere2015best, kiritchenko2016capturing}.
% As the best-worst scaling \cite{louviere2015best} is more reliable than ranking scales \cite{kiritchenko2016capturing}, we compute each system's score as the percentage of times selected as the best minus that is selected as the worst.

Following \citet{bravzinskas2020unsupervised, amplayo2020unsupervised}, we use the following four criteria:
\emph{Fluency}: the summary is grammatically correct, easy to read, and understand;
\emph{Coherence}: the summary is well structured and organized; 
\emph{Informativeness}: the summary mentions specific aspects of the product;
\emph{Redundancy}: the summary has no unnecessary repetitive words or phrases.

\autoref{tbl:humaneval} shows the human evaluation scores of four systems.
In terms of coherence and informativeness, RecurSum achieves the highest score among all approaches across the two datasets.
This result indicates the effectiveness of considering topics and structure in unsupervised abstractive opinion summarization.
As regards fluency, Copycat is superior to our model because our model sometimes makes a grammatical or referential error, which has a negative impact on fluency, as will be shown later in \autoref{sec:analyzesummary}.

\paragraph{Faithfulness of the Summaries}

Abstractive summarization sometimes hallucinates content that is unfaithful to the input texts \cite{maynez2020faithfulness}.
The next study assesses whether the contents mentioned in the generated summaries are included in the input reviews. 
We use the same summary sets as in the quality evaluation and split them into sentences.
For each summary sentence, we asked the AMT workers to judge whether the content is fully mentioned (Full), some of the content is mentioned  (Partial), or no content is mentioned (No) in the reviews.

\autoref{tbl:faithfulness} shows the percentage of each answer. 
The frequency distribution is not regarded as statistically significant by $\chi^2$ test ($p<0.05$).
This result indicates that our model correctly reflects the content in the input reviews as well as Copycat.

\paragraph{Coverage of the Summaries}

Another desirable property of summaries is that they cover more content mentioned in the input reviews.
As reported in \citet{bravzinskas2020unsupervised}, Copycat and MeanSum achieve relatively low scores for the human evaluation of \emph{opinion consensus}, which captures the coverage of common opinions in the input reviews.
% topicを考慮することでcoverageが高くなることを報告している論文があると良い
In contrast, as RecurSum explicitly generates summary sentences for each topic, it could cover more input content across diverse topics.
To assess this assumption, we conducted the opposite study from the faithfulness evaluation.
Similar to the faithfulness evaluation, we split reviews into sentences.
For each review sentence, we asked the AMT workers to rate the extent to which the generated summaries cover the input content. 

\autoref{tbl:coverage} shows the percentage of fully-covered (Full), partially-covered (Partial), and un-covered (No) sentences.
In addition to the two models, we also included gold summaries as the upper bounds.
For both datasets, RecurSum covers more number of common opinions by capturing diverse topics.

\section{Discussion}

\subsection{Analyzing Generated Summaries}
\label{sec:analyzesummary}

In this section, we discuss the strengths and weaknesses of our method by presenting examples of the generated summaries and tree structures.

In \autoref{fig:sample} (a), we present a summary of a review of shoes in Amazon.
RecurSum generates topic sentences about fitness and size (12, 121), similar to Copycat.
In addition, our model also mentions color and use (11, 111, 112), which is also described in the gold summary.
While we cannot grasp that the shoes are appropriate for weddings from Copycat's summary, RecurSum covers such topics and provides more useful information.

\autoref{fig:sample} (b) shows the generated summaries on a coffee shop review in Yelp.
While both RecurSum and Copycat present a positive review about the taste of bubble tea (tea with tapioca), RecurSum also focuses on the dessert (12, 121), similar to the gold summary.
While Copycat also refers to friendly staff, they are not mentioned in the input review. 
Our model successfully does not extract topic sentences about staff by measuring content overlap with the input reviews.
However, RecurSum sometimes makes grammatical or referential errors such as ``\textit{It's a little bit of the best bubble tea}''.
These errors cause the inferior performance of RecurSum in terms of fluency.

\autoref{fig:sample} (c) shows the summary of an Amazon review on a table chair set.
RecurSum accurately captures opinions about the table (11, 111, 112) and chair (12, 121).
The topic sentences on the bottom level elaborate on the parent sentences, referring to the easy assembly (111), the appropriate use of table (112), and the quality of chair (121).
By inferring topics in the tree structure, RecurSum can offer summary sentences over multiple granularities of topics.

\subsection{Ablation Study of Model Components}
\label{sec:ablation}

We report the results of the ablation study to investigate how individual components affect summarization performance.
In addition to the ROUGE scores, we also report self-BLEU scores \cite{zhu2018texygen} to investigate the diversity of the generated summaries.
Self-BLEU is computed by calculating the BLEU score of each generated summary with all other generated summaries in the test set as references.
A higher self-BLEU implies that the generated summaries are not diversified, i.e, the model tends to generate a generic summary similar to the other summaries.
\autoref{tbl:ablation} shows the performances of model variants on Yelp dataset.

\textit{w/o Disc} denotes our model without a discriminator.
The ROUGE scores are significantly lower than the full model.
Without the discriminator, the topic distribution becomes sparse, i.e., most of the review sentences are assigned to some specific topics.
Therefore, the model obtains incoherent topics and generates unfaithful summaries for the input review.
Discriminator penalizes this situation by assigning an appropriate topic to topically different sentences.
This mechanism makes the generated topic sentences topically coherent and improves ROUGE scores.

\textit{w/o Attention} indicates our model without an attention mechanism.
Although the generated sentences are faithful to the input review, they are often generic and miss some specific details of the content.
By adding the attention mechanism, the generated summary effectively reflects the content of the input reviews and provides more detailed information.
Although the copy-mechanism \cite{see2017get} has also been reported to be useful in previous summarization models \cite{bravzinskas2020unsupervised, amplayo2020unsupervised}, it degrades the performance of our model.
While their models use different input-output pairs (reviews vs. pseudo-summary), our model uses the same input-output pairs in an autoencoder manner and tends to fully copy the input sentences.
Thus, our model fails to obtain a meaningful latent code.

\textit{w/o Nucleus} denotes our model using a beam-search decoder (beam width=5) instead of nucleus sampling when decoding topic sentences in inference.
As reported by \citet{holtzman2019curious}, we also confirmed that the beam-search decoder tends to generate bland or repetitive text and sometimes fails to capture product-specific words.
Owing to nucleus sampling, the decoder generates more informative content and improves the ROUGE-1 score with a significant decrease in self-BLEU.

We also attempted to replace the encoder with BERT \cite{devlin2019bert}.
However, fine-tuning of pretrained components with non-pretrained components is unstable as reported by \citet{liu2019text}, and it does not contribute to the improvement of ROUGE scores.

\begin{table}[t!]
\footnotesize
\centering
\begin{tabularx}{\columnwidth}{p{2.3cm}RRR|RR}
\toprule
Model Variants&R-1&R-2&R-L&B-3&B-4\\ 
\midrule
w/o Discriminator&30.52&3.50&16.43&54.18&30.42 \\
w/o Attention&30.62&4.87&17.01&66.11&50.89 \\
w/o Nucleus&31.71&5.10&17.70&69.13&55.81 \\
\midrule
Full&33.24&5.15&18.01&64.30&48.37 \\
\bottomrule
\end{tabularx}
\vspace{-0.5\baselineskip}
\caption{Ablation study of RecurSum on Yelp. R-1/2/L denote ROUGE-1/2/L, respectively. B-3/4 denote self-BLEU3/4, respectively. }
\label{tbl:ablation}
\end{table}

\begin{figure*}[t!]
\begin{minipage}{0.42\hsize}
\centering
\includegraphics[height=4.0cm]{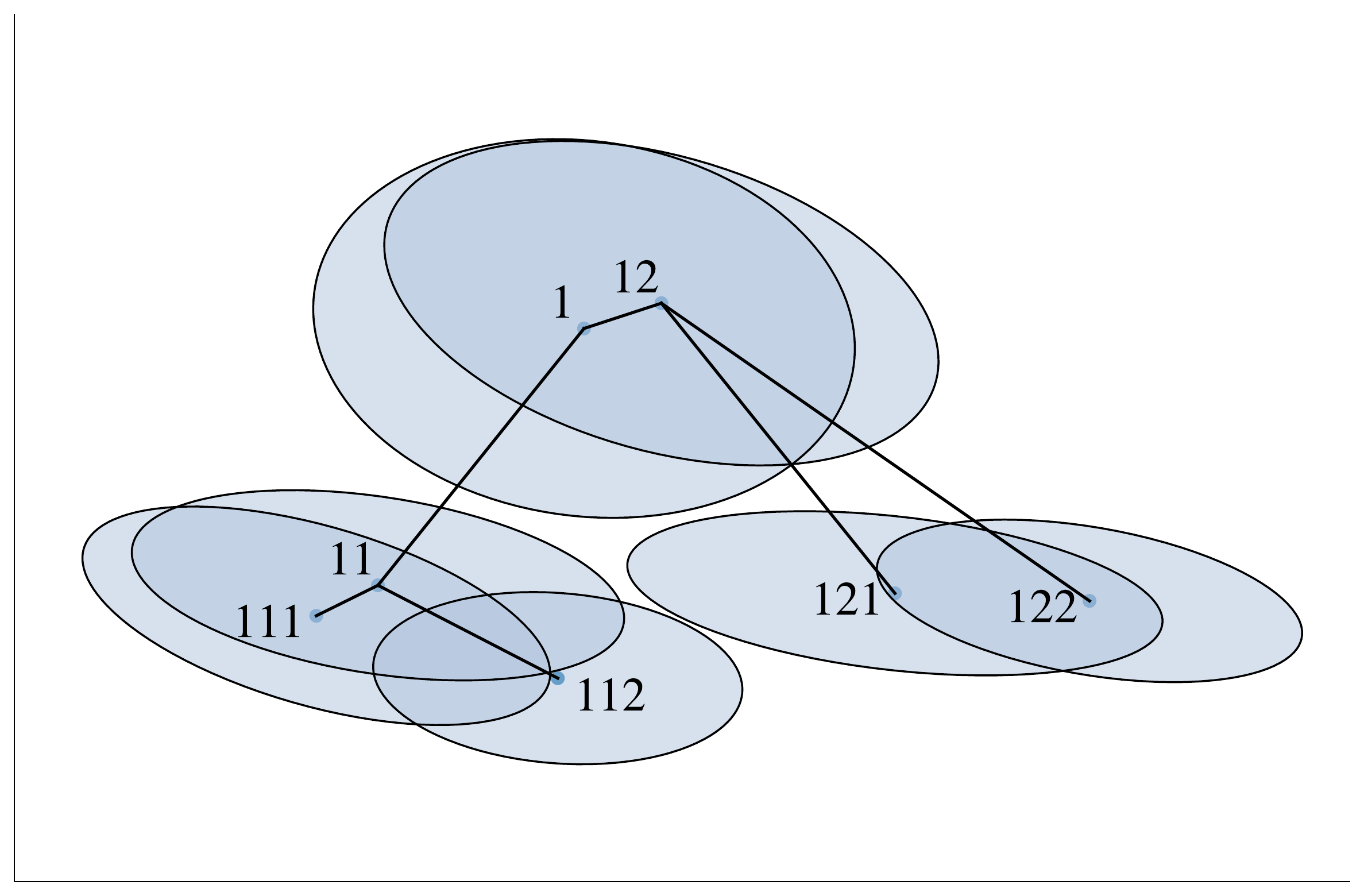}
\end{minipage}
\begin{minipage}{0.56\hsize}
\centering
\includegraphics[height=4.0cm]{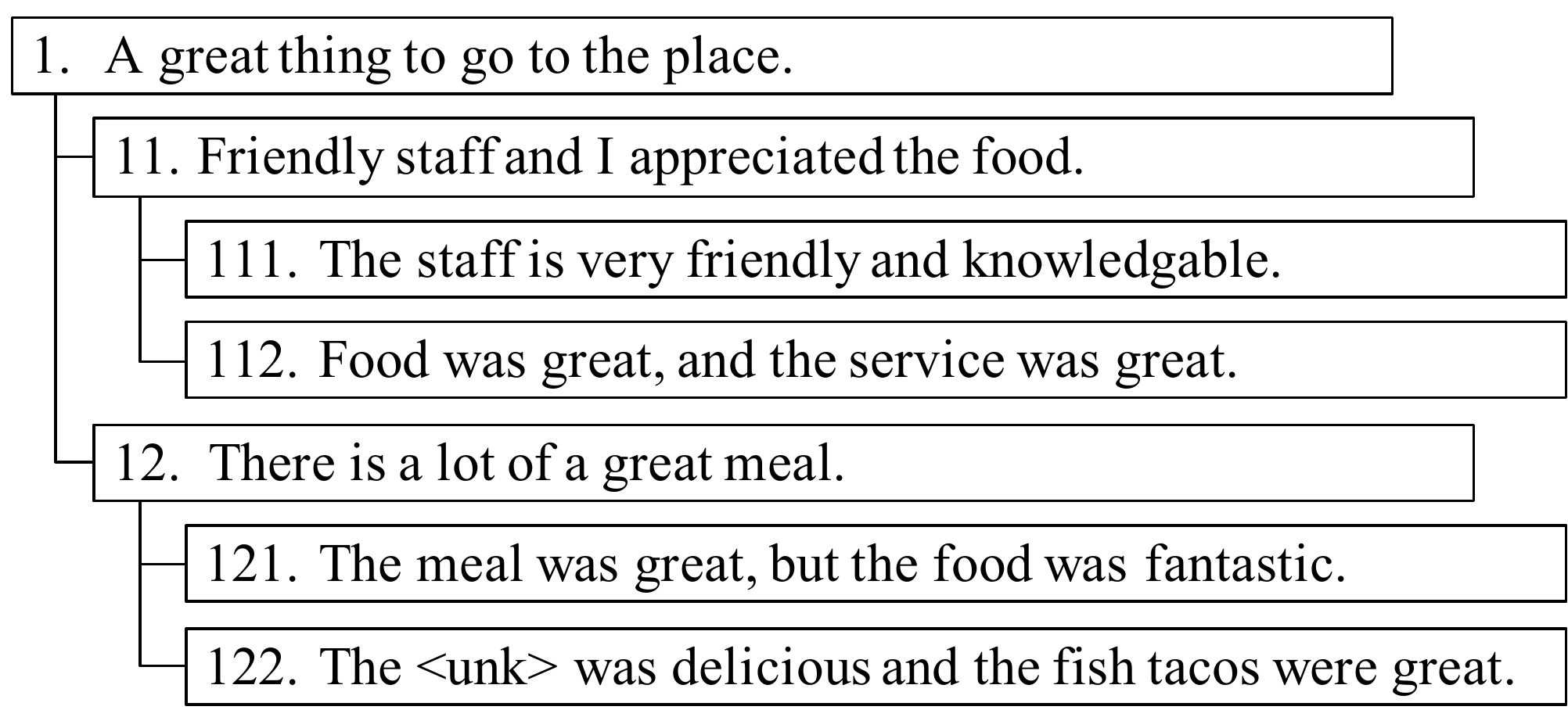}
\end{minipage}
\vspace{-0.5\baselineskip}
\caption{2-D latent space projected by principal component analysis. Each point corresponds to the mean of the latent distribution of a topic sentence, and each circle denotes the same Mahalanobis distance from the mean.}
\vspace{-0.5\baselineskip}
\label{fig:latent}
\end{figure*}

\subsection{Analyzing Topic-Tree Structure}
\label{sec:treestructure}

As generating sentences with tree-structured topic guidance is a novel challenge, we introduce new measures to verify that the generated sentences exhibit the desired properties of tree structures.
Based on the work of tree-structured topic model \cite{kim2012modeling}, we introduce two metrics: \emph{hierarchical affinity} and \emph{topic specialization}.

\paragraph{Hierarchical Affinity:} An important characteristic of the tree structure is that a parent topic sentence is more similar to its children than the sentences descending from the other parents.
To confirm this property, we estimated the similarity of sentences in parent-child pairs and non parent-child pairs.
To measure sentence similarity, we used the \emph{ALBERT} \cite{lan2019albert}, which is a SoTA model on the semantic textual similarity benchmark \cite[\textbf{STS-B};][]{cer2017semeval}.
In our experiment, we used the ALBERT-base, which achieves a $84.7$ Pearson correlation coefficient against the test sets of STS-B.
As shown in \autoref{tbl:hier}, parent-child sentence pairs are more similar than those of non parent-child pairs in both datasets.
This result indicates that the generated sentences linked by parent-child relations are topically coherent.

\paragraph{Topic specialization:} In tree-structured topics, we would expect the root topic to generate general sentences, whereas more specific content is conveyed by the sentences generated by the leaf topics.
To empirically test this property, we estimated the average specificity of sentences at each level of the tree-structured topics.
We fine-tuned ALBERT-base on the task of estimating the specificity of sentences \cite{louis2011automatic}.
We used the dataset provided by \citet{ko2019domain}, which comprises the Yelp, Movie, and Tweet domains.
The fine-tuned model achieves a SoTA performance of $86.2$ Pearson correlation coefficient on the test sets in Yelp.
As shown in \autoref{tbl:spec}, we see that sentences with lower topics are more specific than higher topics.
This indicates that the root sentences refer to general topics, whereas leaf sentences describe more specific topics.

\begin{table}[t!]
\footnotesize
\centering
\begin{tabularx}{\columnwidth}{p{3.6cm}RR}
\toprule
Hierarchical Affinity&Yelp&Amazon \\ 
\midrule
Parent-child pairs&2.39&1.33 \\
Non parent-child pairs&1.59&0.76 \\
\bottomrule
\end{tabularx}
\vspace{-0.5\baselineskip}
\caption{Average sentence similarity of the topic sentence pairs, ranging from 0 (different) to 5 (similar).}
\label{tbl:hier}
\end{table}
\begin{table}[t!]
\footnotesize
\centering
\begin{tabularx}{\columnwidth}{p{3.6cm}RR}
\toprule
Topic Specialization&Yelp&Amazon \\ 
\midrule
First level&1.68&1.60 \\
Second level&1.99&1.63 \\
Third level&2.16&1.84 \\
\bottomrule
\end{tabularx}
\vspace{-0.5\baselineskip}
\caption{Average specialization score of each level topics, ranging from 1 (general) to 5 (specific).}
\label{tbl:spec}
\end{table}

\subsection{Analyzing Latent Space of Sentences}
\label{sec:visualizing}

In \autoref{fig:latent}, we project the latent code of topic sentences of a restaurant review onto the top two principal component vector space.
Following the modelling assumption, the latent distributions of child sentences are located relatively near their parent distributions.
This property ensures that the parent and child sentences are topically coherent, as shown in \autoref{tbl:hier}.
Furthermore, we present the average log determinant of the covariance matrices at each level in \autoref{tbl:logdetcov}.
We confirm that the latent code of the topic sentences has a smaller variance towards the leaves.
This property forces the topic sentences to be more specific as the level becomes deeper, as described in \autoref{tbl:spec}.

\begin{table}[t!]
\footnotesize
\centering
\begin{tabularx}{\columnwidth}{p{3.6cm}RR}
\toprule
LogDetCov&Yelp&Amazon\\ 
\midrule
First level&28.83&27.21 \\ 
Second level&26.22&26.54 \\ 
Third level&23.28&24.55 \\ 
\bottomrule
\end{tabularx}
\vspace{-0.5\baselineskip}
\caption{Average log determinant of covariance matrices (LogDetCov) on each level.}
\label{tbl:logdetcov}
\end{table}

\section{Related Work}

\subsection{Text Generation with Topic Guidance}

The VAE is intensively used to obtain disentangled latent code of sentences \cite{bowman2016generating, hu2017toward, tang2019topic}.
Closely related to ours, \citet{wang2019topic} specify the prior as a GMM, where each mixture component corresponds to the latent code of a topic sentence and is mixed with the topic distribution inferred by the flat neural topic model \cite{miao2017discovering}.

In contrast, we address a novel challenge to generate topic sentences with tree-structured topic guidance, where the root sentence refers to a general topic, whereas the leaf sentences describe more specific topics.
We adopt the tree-structured neural topic model \cite{isonuma2020tree} to infer the topic distribution of sentences and introduce a recursive Gaussian mixture prior for modelling the latent distribution of sentences in a document.

\subsection{Unsupervised Summary Generation}
\label{sec:relatedsummary}

% Most previous work on unsupervised summarization has focused on extractive approaches, such as Lexrank and Opinosis in \autoref{sec:baseline}.
Owing to the success of supervised abstractive summarization by neural architectures \cite{nallapati2016abstractive,see2017get,liu2019hierarchical}, unsupervised sentence compression \cite{fevry2018unsupervised, baziotis2019seq3} and unsupervised summary generation \cite{isonuma2019unsupervised} have recently drawn attention.

Recently, specifically for opinionated texts, several abstractive multi-document summarization methods have been developed, such as MeanSum, Copycat, and DenoiseSum, as explained in \autoref{sec:baseline}.
Concurrently with ours, \citet{angelidis2021extractive} use quantitized transformers enabling aspect-based extractive summarization, and \citet{amplayo2020plansum} incorporate the aspect and sentiment distributions into the unsupervised abstractive summarization.
Our method incorporates topic-tree structure into unsupervised abstractive summarization and generates summaries consisting of multiple granularities of topics.
% These prior unsupervised abstractive approaches generate an opinion summary by capturing consensus opinions from original reviews without considering the topics of each review.
% On the other hand, our model explicitly infers the topics and their tree-structure and generates consensus opinions for each topic.

\section{Conclusion}

In this paper, we proposed a novel unsupervised abstractive opinion summarization method by generating topic sentences with tree-structured topic guidance.
Experimental results demonstrated that the generated summaries are more informative and cover more input content than those generated by the recent unsupervised summarization \cite{bravzinskas2020unsupervised}.
Additionally, we demonstrated that the variance of latent Gaussians represents the granularity of sentences, analogous to Gaussian word embedding \cite{vilnis2015word}.
This property will be useful not only for summarization but also for other tasks that need to consider the granularity of the contents.

\appendix
\section{Appendices}
\label{app:appendices}

\subsection{Inference of Topic Distribution}
\label{app:inferencetopic}

\begin{figure}[t!]
\centering
\includegraphics[width=7.6cm]{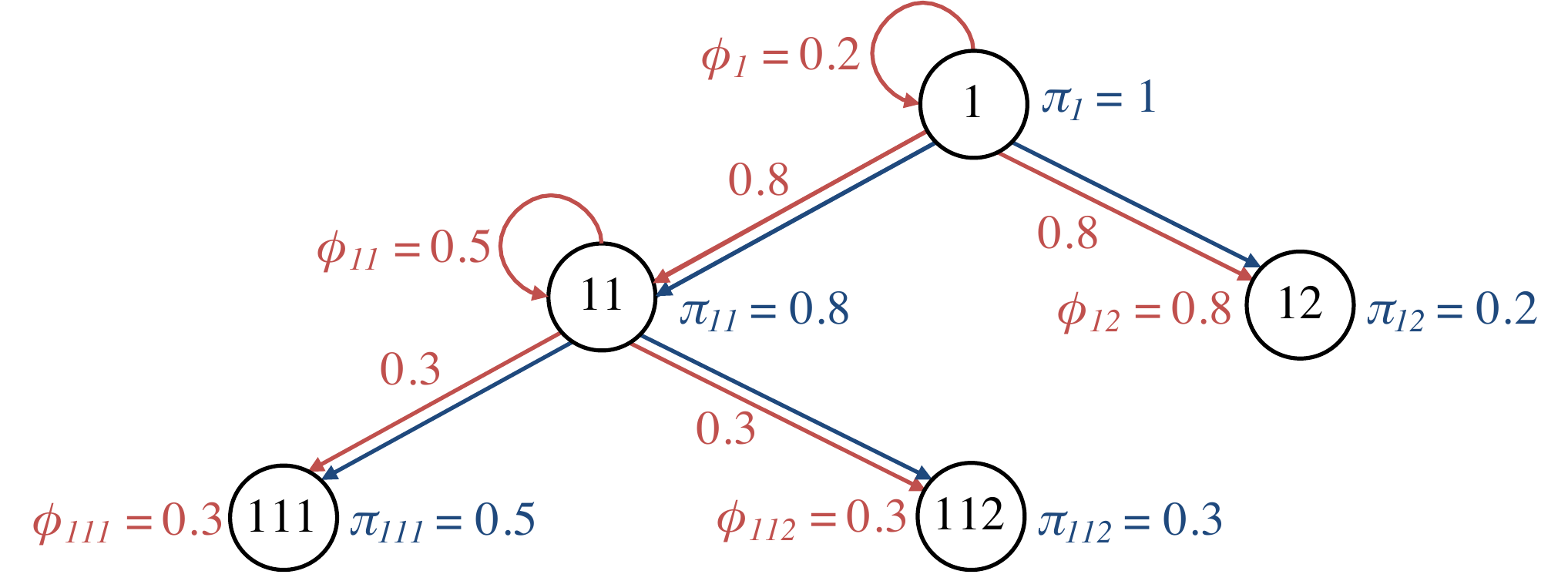}
\vspace{-0.5\baselineskip}
\caption{Example of a path distribution (blue) and level distribution (red). Both the sum of a path distribution over each level and the sum of a level distribution over each path are equal to $1$.}
\label{fig:topicdistribution}
\end{figure}

To approximate the tree-structured topic distribution of a sentence, we use a tree-structured neural topic model \cite[\textbf{TSNTM};][]{isonuma2020tree}, which transforms a sentence into a tree-structured topic distribution using neural networks.
While their model is based on the nested Chinese restaurant process \cite[\textbf{nCRP};][]{griffiths2004hierarchical}, we make a minor change to use the nested hierarchical Dirichlet process \cite[\textbf{nHDP};][]{paisley2014nested}.
The nHDP generates a sentence-specific path distribution $\bm{\pi}_s$ and level distribution $\bm{\phi}_s$ as

{\footnotesize
\vspace{-\baselineskip}
\begin{align}
\nu_{s,k} \!\sim\! \mathrm{Beta}(1, \gamma), \
&\pi_{s,k} \!=\! \pi_{s,par(k)} \nu_{s,k} \prod_{j \in \mathrm{Sib}(k)}(1-\nu_{s,j}) \label{eq:tsbp} \\
\eta_{s,k} \!\sim\! \mathrm{Beta}(\alpha, \beta), \
&\phi_{s,k} \!=\! \eta_{s,k} \prod_{j \in \mathrm{Anc}(k)}(1-\eta_{s,j}) \label{eq:sbp} \\
\theta_{s,k} \!=\! \pi_{s,k} \cdot \phi_{s,k} \label{eq:nhdp}
\end{align}
}%
where $\mathrm{Sib}(k)$ and $\mathrm{Anc}(k)$ are the sets of the $k$-th topic's preceding-siblings and ancestors, respectively.
As described in \autoref{fig:topicdistribution}, $\pi_{s,k}$ denotes the probability that a sentence $s$ selects a path from the root to the $k$-th topic.
$\phi_{s,k}$ denotes the probability that a sentence $s$ does not select the ancestral topics $j \in \mathrm{Anc}(k)$ but remains in the $k$-th topic along the path.
By multiplying these two probabilities, we obtain $\theta_{s,k}$; the probability that a sentence $s$ selects the topic $k$.
The nHDP does not make a significant difference in the summarization performance from the nCRP.
However, the nHDP permits different lengths of each path, whereas the nCRP restricts each path length to be the same.

Following \citet{isonuma2020tree}, we use the doubly-recurrent neural networks \cite[\textbf{DRNN};][]{alvarez2017tree} to transform a sentence embedding $\bm{y}_s\!=\!\mathrm{RNN}(\bm{w}_s)$ to the path distribution $\bm{\pi}_s$ and level distribution $\bm{\phi}_s$.
The DRNN consists of two RNN decoders over respectively the ancestors and siblings.
We compute the $k$-th topic's hidden state $\bm{h}_k$ using \eqref{eq:drnn} and obtain the path distribution by alternating $\bm{\nu}_s$ as \eqref{eq:breakingproportion}:

{\small
\vspace{-\baselineskip}
\begin{align}
\label{eq:drnn}
\bm{h}_k &= \mathrm{tanh}(\bm{W}_p\bm{h}_{par(k)} + \bm{W}_s\bm{h}_{k-1}) \\
\nu_{s,k} &= \mathrm{sigmoid}(\bm{h}_k^\top \bm{y}_s) \label{eq:breakingproportion}
\end{align}
}%
where $\bm{h}_{par(k)}$ and $\bm{h}_{k-1}$ are the hidden states of a parent and a previous sibling of the $k$-th topic, respectively. 
Similarly, we obtain the level distribution, $\bm{\phi}_s$, by computing $\bm{\eta}_s$ with another DRNN.

\begin{table}[t!]
\footnotesize
\centering
\begin{tabularx}{\columnwidth}{p{4cm}RRR}
\toprule
\# of topics for each level (total) &R-1&R-2&R-L\\ 
\midrule
1--2--4 (7)&29.03&4.39&16.94 \\
1--3--9 (13)&31.42&4.43&17.19 \\
1--4--16 (21)&33.24&5.15&18.01 \\
1--5--25 (31)&31.94&4.78&17.50 \\
1--6--36 (43)&33.25&4.82&17.81 \\
\bottomrule
\end{tabularx}
\vspace{-0.5\baselineskip}
\caption{Sensitivity for various number of branches.}
\label{tbl:sensitivity_branch}
\end{table}

\begin{table}[t!]
\footnotesize
\centering
\begin{tabularx}{\columnwidth}{p{4cm}RRR}
\toprule
\# of topics for each level (total) &R-1&R-2&R-L\\ 
\midrule
1--3 (4)&23.63&2.38&14.35 \\
1--3--9 (13)&31.42&4.43&17.19 \\
1--3--9--27 (40)&32.55&4.75&17.70 \\
\bottomrule
\end{tabularx}
\vspace{-0.5\baselineskip}
\caption{Sensitivity for various number of levels.}
\label{tbl:sensitivity_level}
\end{table}

\subsection{Sensitivity for the Number of Topics}

We investigated how the number of topics affects summarization performance.
\autoref{tbl:sensitivity_branch} shows the ROUGE scores on the various number of branches with a fixed depth of $3$ in topic-tree structure.
When the number of topics is small, the models achieve a relatively low score.
However, when the number of branches $\geq 4$, the performance does not significantly change for various numbers of topics.
A similar trend is confirmed in \autoref{tbl:sensitivity_branch}, which shows the ROUGE scores on the various number of levels with the fixed number of branches of $3$.
These results indicate that our model is relatively robust for the number of topics.

\subsection{Derivation of the Equation \eqref{eq:kl_geq}}
\label{app:proof}

\paragraph{Proposition:} when $q(\bm{x}_s|z_s)$ is given by \eqref{eq:em}, \eqref{eq:proposition} holds:

{\footnotesize
\vspace{-\baselineskip}
\begin{align}
\begin{split}
\label{eq:proposition}
&{\textstyle \sum_s \hat{\theta}_{s,k} \mathrm{D_{KL}}\bigl[q(\bm{x}_s|\bm{w}_s)|p(\bm{x}_s|z_s\!=\!k) \bigr] - }\\
&{\textstyle \!\-\! \sum_s \hat{\theta}_{s,k} \mathrm{D_{KL}}\bigl[q(\bm{x}_s|z_s\!=\!k)|p(\bm{x}_s|z_s\!=\!k) \bigr] \geq 0}
\end{split}
\end{align}
}%

\paragraph{Proof:} The first term of \eqref{eq:proposition} is re-written as:

{\footnotesize
\vspace{-\baselineskip}
\begin{align}
\begin{split}
&{\textstyle \sum_s \hat{\theta}_{s,k} \mathrm{D_{KL}}\bigl[q(\bm{x}_s|\bm{w}_s)|p(\bm{x}_s|z_s\!=\!k) \bigr]}\\
=&{\textstyle \sum_s \hat{\theta}_{s,k} \mathrm{D_{KL}}\bigl[\mathcal{N}(\hat{\bm{\mu}}_s, \hat{\bm{\Sigma}}_s)|\mathcal{N}(\hat{\bm{\mu}}_{d,par(k)}, \hat{\bm{\Sigma}}_{d,par(k)}) \bigr] }\\
=& \frac{1}{2} {\textstyle \sum_s \hat{\theta}_{s,k} \Bigl\{\log|\hat{\bm{\Sigma}}_{d,par(k)}| \!-\! \log|\hat{\bm{\Sigma}}_s| \!+\! \mathrm{Tr}[\hat{\bm{\Sigma}}_{d,par(k)}^{-1}\hat{\bm{\Sigma}}_s]} \\
    &\!+\! {\textstyle (\hat{\bm{\mu}}_s \!-\! \hat{\bm{\mu}}_{d,par(k)})^\top \hat{\bm{\Sigma}}_{d,par(k)}^{-1} (\hat{\bm{\mu}}_s \!-\! \hat{\bm{\mu}}_{d,par(k)}) \!-\! n \Bigr\}}\\
=& \frac{1}{2} {\textstyle \sum_s \hat{\theta}_{s,k} \Bigl\{C_{d,par(k)} \!-\! \log|\hat{\bm{\Sigma}}_s| \!+\! \mathrm{Tr}[\hat{\bm{\Sigma}}_{d,par(k)}^{-1}\hat{\bm{\Sigma}}_s]} \\
    &\!+\! {\textstyle \hat{\bm{\mu}}_s^\top \hat{\bm{\Sigma}}_{d,par(k)}^{-1} \hat{\bm{\mu}}_s \!-\! 2\hat{\bm{\mu}}_{d,par(k)}^\top \hat{\bm{\Sigma}}_{d,par(k)}^{-1} \hat{\bm{\mu}}_s \Bigr\}}\\
=& \frac{1}{2} {\textstyle \sum_s \hat{\theta}_{s,k} \Bigl\{C_{d,par(k)} \!-\! \log|\hat{\bm{\Sigma}}_s| \!+\! \mathrm{Tr}[\hat{\bm{\Sigma}}_{d,par(k)}^{-1}\hat{\bm{\Sigma}}_s]}\\
    &\!+\! {\textstyle \mathrm{Tr}[\hat{\bm{\Sigma}}_{d,par(k)}^{-1} \hat{\bm{\mu}}_s \hat{\bm{\mu}}_s^\top] \!-\! 2\hat{\bm{\mu}}_{d,par(k)}^\top \hat{\bm{\Sigma}}_{d,par(k)}^{-1} \hat{\bm{\mu}}_{d,k} \Bigr\}}
\end{split}
\end{align}
}%
as $\sum_s \hat{\theta}_{s,k} \hat{\bm{\mu}}_{d,k} \!=\! \sum_s \hat{\theta}_{s,k} \hat{\bm{\mu}}_s$ from \eqref{eq:mu_k}.

\newpage
The second term is similarly expanded as:

\vspace{-10pt}
\footnotesize
\begin{align}
\begin{split}
& {\textstyle \sum_s \hat{\theta}_{s,k} \mathrm{D_{KL}}\bigl[q(\bm{x}_s|z_s\!=\!k)|p(\bm{x}_s|z_s\!=\!k) \bigr]} \\
=& \frac{1}{2} {\textstyle \sum_s \hat{\theta}_{s,k} \Bigl\{C_{d,par(k)} \!-\! \log|\hat{\bm{\Sigma}}_{d,k}| \!+\! \mathrm{Tr}[\hat{\bm{\Sigma}}_{d,par(k)}^{-1}\hat{\bm{\Sigma}}_{d,k}]} \\
    &\!+\! {\textstyle \mathrm{Tr}[\hat{\bm{\Sigma}}_{d,par(k)}^{-1} \hat{\bm{\mu}}_{d,k} \hat{\bm{\mu}}_{d,k}^\top] \!-\! 2\hat{\bm{\mu}}_{d,par(k)}^\top \hat{\bm{\Sigma}}_{d,par(k)}^{-1} \hat{\bm{\mu}}_{d,k} \Bigr\}}
\end{split}
\end{align}
\normalsize

Therefore, \eqref{eq:proposition} is arranged as:
\footnotesize
\begin{align}
\begin{split}
&{\textstyle \sum_s \hat{\theta}_{s,k} \mathrm{D_{KL}}\bigl[q(\bm{x}_s|\bm{w}_s)|p(\bm{x}_s|z_s\!=\!k) \bigr]} \\ 
&\!-\! {\textstyle \sum_s \hat{\theta}_{s,k} \mathrm{D_{KL}}\bigl[q(\bm{x}_s|z_s\!=\!k)|p(\bm{x}_s|z_s\!=\!k) \bigr]} \\
=& \frac{1}{2} {\textstyle \sum_s \hat{\theta}_{s,k} \Bigl\{\!-\!\log|\hat{\bm{\Sigma}}_s| \!+\! \log|\hat{\bm{\Sigma}}_{d,k}|} \\
&\!+\! {\textstyle \mathrm{Tr} \bigl[\hat{\bm{\Sigma}}_{d,par(k)}^{-1} (\hat{\bm{\Sigma}}_s \!+\! \hat{\bm{\mu}}_s \hat{\bm{\mu}}_s^\top \!-\! \hat{\bm{\Sigma}}_{d,k} \!-\! \hat{\bm{\mu}}_{d,k} \hat{\bm{\mu}}_{d,k}^\top) \bigr] \Bigr\}} \\
=& \frac{1}{2} {\textstyle \sum_s \hat{\theta}_{s,k} \Bigl\{\!-\!\log|\hat{\bm{\Sigma}}_s| \!+\! \log|\hat{\bm{\Sigma}}_{d,k}| \Bigr\}} \!+\! \frac{1}{2} {\textstyle \Bigl\{ \mathrm{Tr} \bigl[\hat{\bm{\Sigma}}_{d,par(k)}^{-1}} \\
&{\textstyle \sum_s \hat{\theta}_{s,k}  (\hat{\bm{\Sigma}}_s \!+\! \hat{\bm{\mu}}_s \hat{\bm{\mu}}_s^\top \!-\! \hat{\bm{\Sigma}}_{d,k} \!-\! \hat{\bm{\mu}}_{d,k} \hat{\bm{\mu}}_{d,k}^\top) \bigr] \Bigr\}} \\
=& \frac{1}{2} {\textstyle \sum_s \hat{\theta}_{s,k} \Bigl\{-\log|\hat{\bm{\Sigma}}_s| + \log|\hat{\bm{\Sigma}}_{d,k}| \Bigr\}} \label{eq:entropy}
\end{split}
\end{align}
\normalsize
as $\sum_s \hat{\theta}_{s,k} \bigl\{\hat{\bm{\Sigma}}_s \!+\! \hat{\bm{\mu}}_s \hat{\bm{\mu}}_s^\top \bigr\} = \sum_s \hat{\theta}_{s,k} \bigl\{ \hat{\bm{\Sigma}}_{d,k} \!+\! \hat{\bm{\mu}}_{d,k} \hat{\bm{\mu}}_{d,k}^\top \bigr\}$ from \eqref{eq:Sigma_k}. The given equation eventually comes down to a comparison of the entropy.

Since, in general,  $\!-\! \int q_1(\bm{x}) \log q_2(\bm{x}) \ d\bm{x} \!\geq\! \!-\!\int q_1(\bm{x}) \log q_1(\bm{x}) \ d\bm{x}$ holds, we obtain \eqref{eq:crossentropy}:
\footnotesize
\begin{align}
\begin{split}
{\textstyle \sum_s \hat{\theta}_{s,k} \Bigl\{- \int q(\bm{x}_s|\bm{w}_s) \log q(\bm{x}_s|z_s\!=\!k) \ d\bm{x_s} \Bigr\}} \\
\!\geq\! {\textstyle \sum_s \hat{\theta}_{s,k} \Bigl\{- \int q(\bm{x}_s|\bm{w}_s) \log q(\bm{x}_s|\bm{w}_s) \ d\bm{x_s} \Bigr\}}
\label{eq:crossentropy}
\end{split}
\end{align}
\normalsize

As the right term is a weighted sum of the normal distribution entropy, it can be rewritten as:
\footnotesize
\begin{align}
\begin{split}
&{\textstyle \sum_s \hat{\theta}_{s,k} \Bigl\{\!-\! \int q(\bm{x}_s|\bm{w}_s) \log q(\bm{x}_s|\bm{w}_s) \ d\bm{x_s} \Bigr\}} \\
=& \frac{1}{2} {\textstyle \sum_s \hat{\theta}_{s,k} \Bigl\{\log|\hat{\bm{\Sigma}}_s| \!+\! n\log 2\pi + n \Bigr\}} \label{eq:right}
\end{split}
\end{align}
\normalsize

Meanwhile, we can expand the left term as:
\footnotesize
\begin{align}
\begin{split}
&{\textstyle \sum_s \hat{\theta}_{s,k} \Bigl\{\!-\! \int q(\bm{x}_s|z_s\!=\!k) \log q(\bm{x}_s|\bm{w}_s) \ d\bm{x_s} \Bigr\}} \\
=& \frac{1}{2} {\textstyle \sum_s \hat{\theta}_{s,k} \Bigl\{\log|\hat{\bm{\Sigma}}_{d,k}| \!+\! n\log 2\pi} \\
&\!+\! {\textstyle \mathrm{E}_{q(\bm{x}_s|\bm{w}_s)}[(\bm{x}_s\!-\!\hat{\bm{\mu}}_{d,k})^\top \hat{\bm{\Sigma}}_{d,k}^{-1} (\bm{x}_s\!-\!\hat{\bm{\mu}}_{d,k})] \Bigr\}} \label{eq:left}
\end{split}
\end{align}
\normalsize

The last term in \eqref{eq:left} is expressed as:
\footnotesize
\begin{align}
\begin{split}
& {\textstyle \sum_s \hat{\theta}_{s,k} \Bigl\{ \mathrm{E}_{q(\bm{x}_s|\bm{w}_s)} \bigl[ (\bm{x}_s\!-\!\hat{\bm{\mu}}_{d,k})^\top \hat{\bm{\Sigma}}_{d,k}^{-1} (\bm{x}_s\!-\!\hat{\bm{\mu}}_{d,k}) \bigr] \Bigr\}} \\
=& {\textstyle \sum_s \hat{\theta}_{s,k} \Bigl\{ \mathrm{E}_{q(\bm{x}_s|\bm{w}_s)}[\mathrm{Tr}(\hat{\bm{\Sigma}}_{d,k}^{-1} (\bm{x}_s\!-\!\hat{\bm{\mu}}_{d,k}) (\bm{x}_s\!-\!\hat{\bm{\mu}}_{d,k})^\top) ] \Bigr\}} \\
=& {\textstyle \mathrm{Tr}\Bigl[ \hat{\bm{\Sigma}}_{d,k}^{-1} \sum_s \hat{\theta}_{s,k} \bigl\{ \mathrm{E}_{q(\bm{x}_s|\bm{w}_s)}[(\bm{x}_s\!-\!\hat{\bm{\mu}}_{d,k}) (\bm{x}_s\!-\!\hat{\bm{\mu}}_{d,k})^\top] \bigr\} \Bigr]} \\
=& {\textstyle \mathrm{Tr}\Bigl[ \hat{\bm{\Sigma}}_{d,k}^{-1} \sum_s \hat{\theta}_{s,k} \hat{\bm{\Sigma}}_{d,k} \Bigr]} \\
=& {\textstyle \sum_s \hat{\theta}_{s,k} \ n} \label{eq:leftlast}
\end{split}
\end{align}
\normalsize

Thus, by combining \eqref{eq:crossentropy}, \eqref{eq:right}, \eqref{eq:left}, \eqref{eq:leftlast}, $\sum_s \hat{\theta}_{s,k} \bigl\{\log|\hat{\bm{\Sigma}}_{d,k}| \bigr\} \!\geq\! \sum_s \hat{\theta}_{s,k} \bigl\{\log|\hat{\bm{\Sigma}}_s|\bigr\}$ holds and implies \eqref{eq:proposition}.

\section*{Acknowledgments}

We would like to thank the anonymous reviewers and action editor, Asli Celikyilmaz, for their valuable feedback.
This work was supported by JST ACT-X Grant Number JPMJAX1904 and JSPS KAKENHI Grant Number JP20J10726, Japan.

\bibliography{tacl2018}
\bibliographystyle{acl_natbib}

\end{document}